\documentclass[conference]{IEEEtran}

\ifCLASSOPTIONcompsoc
	\usepackage[nocompress]{cite}
\else
	\usepackage{cite}
\fi

\usepackage{amsmath,amssymb,amsfonts}
\usepackage{filecontents}
\usepackage{algorithmic}
\usepackage{graphicx}
\usepackage{subcaption}
\usepackage{textcomp}
\usepackage{xcolor}
\usepackage{enumitem}
\usepackage{calc}
\usepackage{tikz}
\usepackage{tikz-3dplot}
\usepackage{multirow}
\usepackage{mathtools}
\usepackage{pifont}
\usepackage[hidelinks, urlcolor=green]{hyperref}

\hypersetup{
    colorlinks,
	citecolor={green},
	urlcolor={black}
}

\DeclarePairedDelimiter\floor{\lfloor}{\rfloor}

\def\BibTeX{{\rm B\kern-.05em{\sc i\kern-.025em b}\kern-.08em
    T\kern-.1667em\lower.7ex\hbox{E}\kern-.125emX}}
    
\graphicspath{{./images/}}
\usetikzlibrary{calc,arrows.meta,positioning,backgrounds}

\newcommand{\doubleimage}[2]{\begin{figure}[h!] \centering \begin{subfigure}[b]{0.475\linewidth} \includegraphics[width=\linewidth]{#1} \end{subfigure} \begin{subfigure}[b]{0.05\linewidth} \end{subfigure} \begin{subfigure}[b]{0.475\linewidth} \includegraphics[width=\linewidth]{#2} \end{subfigure} \end{figure}}

\begin{document}

\title{Real-Time Freespace Segmentation on Autonomous Robots for Detection of Obstacles and Drop-Offs}

\author{\IEEEauthorblockN{Anish Singhani\\research@anishsinghani.com}}

\maketitle

\begin{abstract}
Mobile robots navigating in indoor and outdoor environments must be able to identify and avoid unsafe terrain. Although a significant amount of work has been done on the detection of standing obstacles (solid obstructions), not much work has been done on the detection of negative obstacles (e.g. dropoffs, ledges, downward stairs). We propose a method of terrain safety segmentation using deep convolutional networks. Our custom semantic segmentation architecture uses a single camera as input and creates a freespace map distinguishing safe terrain and obstacles. We then show how this freespace map can be used for real-time navigation on an indoor robot. The results show that our system generalizes well, is suitable for real-time operation, and runs at around 55 fps on a small indoor robot powered by a low-power embedded GPU.
\end{abstract}

\section{Introduction}

Many small unmanned ground robots are being developed to perform tasks such as delivery, surveillance, household tasks, etc. These robots must navigate difficult terrain, requiring advanced perception capabilities. These robots use various sensors, including 2D and 3D LIDAR, RGB cameras, radar, ultrasonic, infrared sensors, and depth-sensing cameras. Most of these sensors, excluding LIDAR and RGB cameras, suffer from low resolution and/or short range. For this reason, most robots use a 2D or 3D LIDAR sensor for mapping and real-time obstacle avoidance.

LIDAR sensors use a scanning laser time-of-flight sensor to estimate distance extremely precisely. 2D (planar) LIDAR has high resolution and range, but can only detect opaque surfaces that intersect the plane made by the sweeping LIDAR beam. A lot of the work in the robotics field \cite{cite:21} \cite{cite:22} has focused on detection of obstacles using laser and visual sensors, but less work has been done on detecting drop-offs and other hazards, frequently referred to as negative obstacles \cite{cite:23}. These include ledges, staircases, and raised curbs, which are all examples of the absence of traversable ground. Most methods of detecting negative obstacles \cite{cite:24} require 3D LIDAR sensor. Although it can successfully can detect these obstacles, it is extremely costly. When used on outdoor robots, LIDAR sensors are also very sensitive to light refraction due to rainwater.

In this paper we use RGB cameras instead of laser-based sensors. These cameras have several advantages. They are cheap and small enough that they can be used on small, low-cost robots. For a much lower cost than a single LIDAR, it is possible to mount several cameras at different angles, providing a wider field of view \cite{cite:26} and redundancy, both of which are important for safety. Additionally, cameras are much more versatile. They can be used for many purposes, including object detection \cite{cite:25} and remote monitoring.

Using Convolutional Neural Networks, it is possible to detect and segment obstacles from free space in an image, to allow the robot to safely avoid obstacles. We define free space as regions in a camera image corresponding to terrain which a robot can safely drive on. Our approach aims to detect all types of obstacles, including those which cannot be detected by LIDAR: negative obstacles, ledges, overhangs, glass surfaces, etc.

\begin{figure}
	\includegraphics[width=\linewidth]{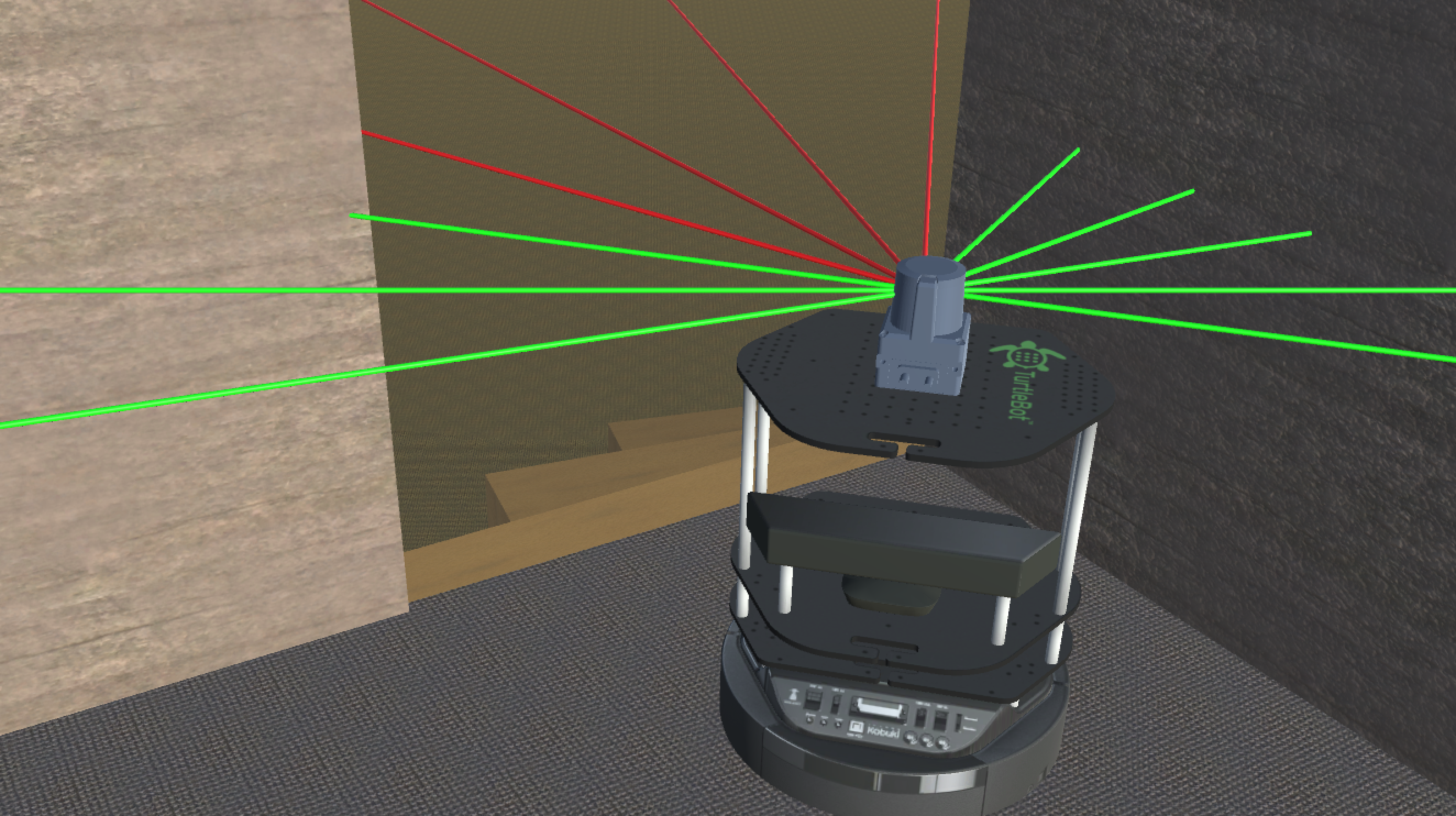}
	\caption{While 2D laser-based sensors can detect standing obstacles (such as walls) very accurately (green laser beams), they fail to detect negative obstacles such as the downward staircase pictured behind the robot (red laser beams).}
	\label{fig:turtlebot_stairs}
\end{figure}

\section{Related Work}

In the context of indoor and outdoor robot navigation, most development has been based on the use of LIDAR or depth-sensing cameras for localization and avoidance \cite{cite:14} \cite{cite:16}. These approaches create a pointcloud from data captured by the LIDAR or depth camera. This pointcloud is then used to build a costmap \cite{cite:15}, which stores locations of obstacles and free space. Pathfinding algorithms can be used on this costmap to find a safe path for a robot to travel.

Segmentation has conventionally been approached in two different ways: instance segmentation \cite{cite:28} and pixel-wise semantic segmentation \cite{cite:8}. In instance segmentation, each object in an image is isolated and both a segmentation overlay and a bounding box is calculated. In pixel-wise segmentation, every pixel in an image is labeled with a classification. For our use-case of free space detection, we focus on pixel-wise semantic segmentation.

There are two main approaches to pixel-wise segmentation: encoder-decoder CNNs \cite{cite:8} and atrous convolution with bilinear upsampling \cite{cite:19}. The former uses convolutions and pooling layers to downsample an input image, then uses learned deconvolutions \cite{cite:29} and/or unpooling layers, along with skip connections to create an output image with the same resolution as the input and each pixel labeled with a classification. The latter uses atrous (dilated) convolutions to increase the effective receptive field of the convolutions while downsampling less than the encoder-decoder architecture, and uses simple bilinear upsampling on the output classifications to resize the output image to the same size as the input. Due to the atrous convolution's larger receptive field and ability to parse features at different scales, it has been much more successful at general-purpose semantic segmentation tasks than encoder-decoder architectures.

\section{Approach}

The main goal of our work is to allow an indoor robot to accurately detect free space using an RGB camera. We apply an end-to-end semantic segmentation model developed to be accurate while running in real-time on low-power embedded hardware. We then integrate the output of the segmentation into a real-time autonomous navigation stack running on an indoor robot.

Our pipeline is separated into three steps. First, a frame is captured from the camera and simple edge detection techniques are applied. Next, the frame and detected edges are processed by our semantic segmentation neural network to produce a freespace map. Finally, this freespace map is converted to a 3D pointcloud, which is added to a navigation map to allow real-time navigation.

\subsection{Input Preprocessing}

Segmentation neural networks are generally trained with minimal domain-specific data preprocessing, to allow the optimizer to learn the best way to interpret the data by analyzing large amounts of labeled training data. However, this requires a large amount of labeled training data to ensure the learned parameters are robust to variations in the environment. In the case of semantic segmentation, the cost of labeling more images is high, a result of the precise nature of the necessary annotations. 

To alleviate this problem and speed up the neural network's convergence, we add additional preprocessed channels to the network's input. These allow the network to more easily learn to refine edges of obstacles in the segmentation output by adding information about the target domain (i.e. physical edges of obstacles are generally accompanied by variations in color). 

In order to calculate regions that are likely to be physical edges of obstacles, we use a form of edge detection based on computing the gradient of the image \cite{cite:12}. We compute the discrete Sobel \cite{cite:13} (1st order derivative) and Laplacian (2nd order derivative) gradients of the greyscale of the input image. These three gradients (Sobel X, Sobel Y, and Laplacian) are then combined into an image tensor, downsampled to 112x112 (half the input size), and used as an auxiliary input to the segmentation neural network.

%\begin{align*}
%S_x(I) &= \left(\begin{IEEEeqnarraybox*}[][c]{,c/c/c,} -1&+0&+1\\ -2&+0&+2\\ -1&+0&+1 \end{IEEEeqnarraybox*}\right) \ast I \\ \\
%S_y(I) &= \left(\begin{IEEEeqnarraybox*}[][c]{,c/c/c,} -1&-2&-1\\ +0&+0&+0\\ +1&+2&+1 \end{IEEEeqnarraybox*}\right) \ast I \\ \\
%L(I) &= \left(\begin{IEEEeqnarraybox*}[][c]{,c/c/c,} +0&-1&+0\\ -1&+4&-1\\ +0&-1&+0 \end{IEEEeqnarraybox*}\right) \ast I
%\end{align*}

%\begin{table}[h!]
%	\renewcommand{\arraystretch}{1.3}
%	\caption{}
%	\centering
%	\begin{tabular}{l|l|l}
%	\hline
%	\multicolumn{3}{l}{Input Channels for CNN} \\ \hline
%	\multirow{3}{*}{Main Input} & 1       & Red             \\ 
%								& 2       & Green           \\ 
%								& 3       & Blue            \\ \hline
%	\multirow{3}{*}{Aux Input}  & 1       & Sobel X         \\ 
%								& 2       & Sobel Y         \\ 
%								& 3       & Laplacian       \\ \hline
%	\end{tabular}
%\end{table}

\begin{figure}[h!]
	\centering
	\begin{subfigure}[b]{0.45\linewidth}
	  \includegraphics[width=\linewidth]{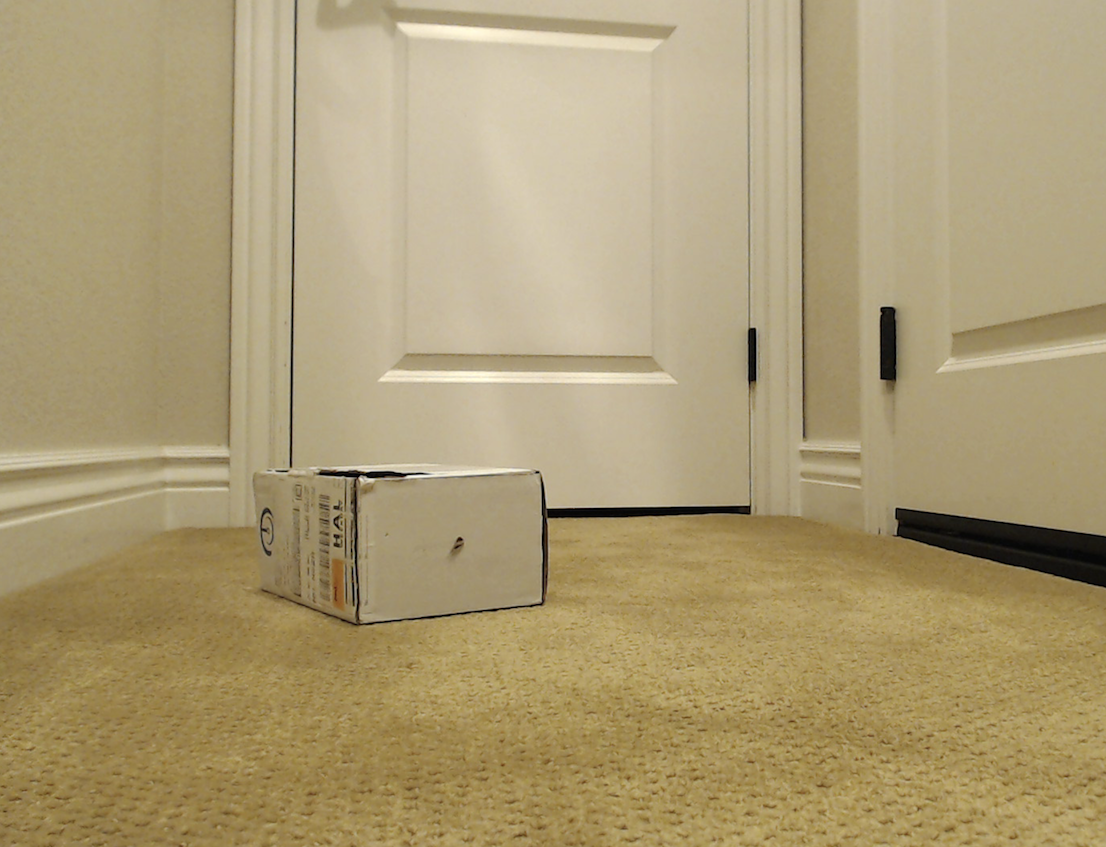}
	  \caption{Original RGB}
	\end{subfigure}
	\begin{subfigure}[b]{0.45\linewidth}
	  \includegraphics[width=\linewidth]{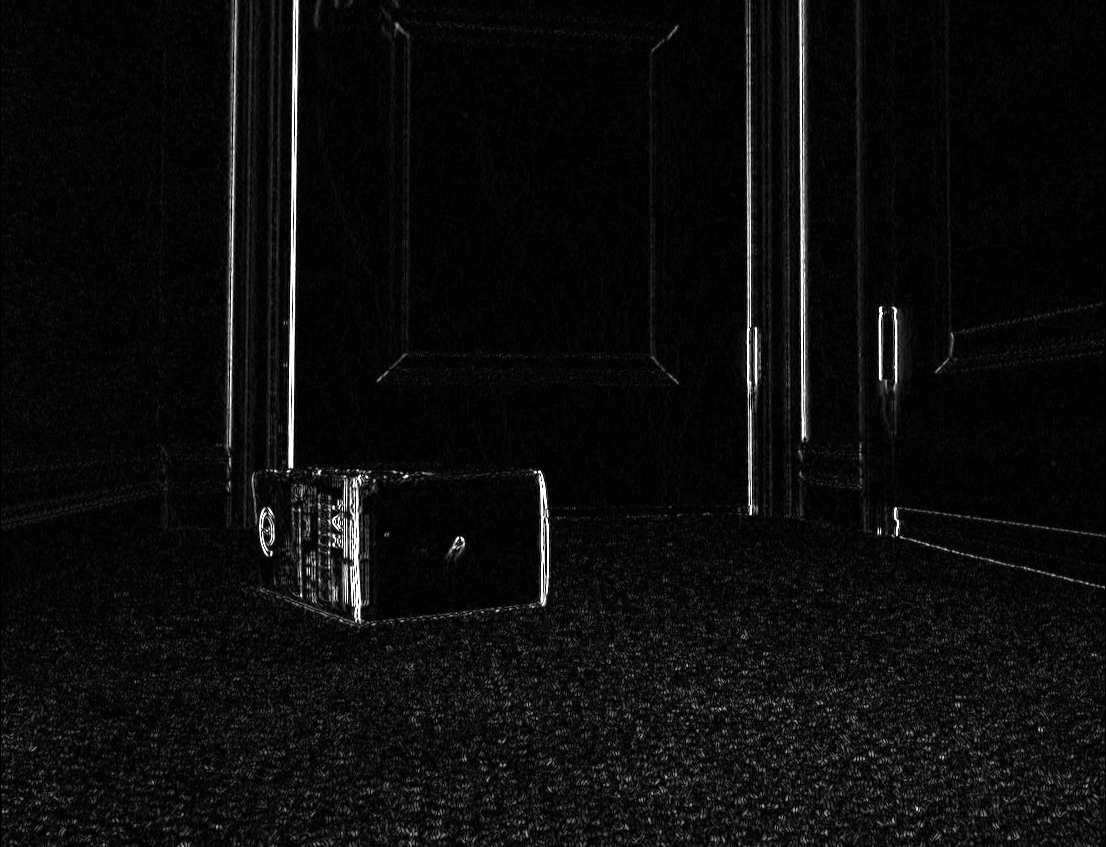}
	  \caption{Sobel X}
	\end{subfigure}
	\par\bigskip 
	\begin{subfigure}[b]{0.45\linewidth}
	  \includegraphics[width=\linewidth]{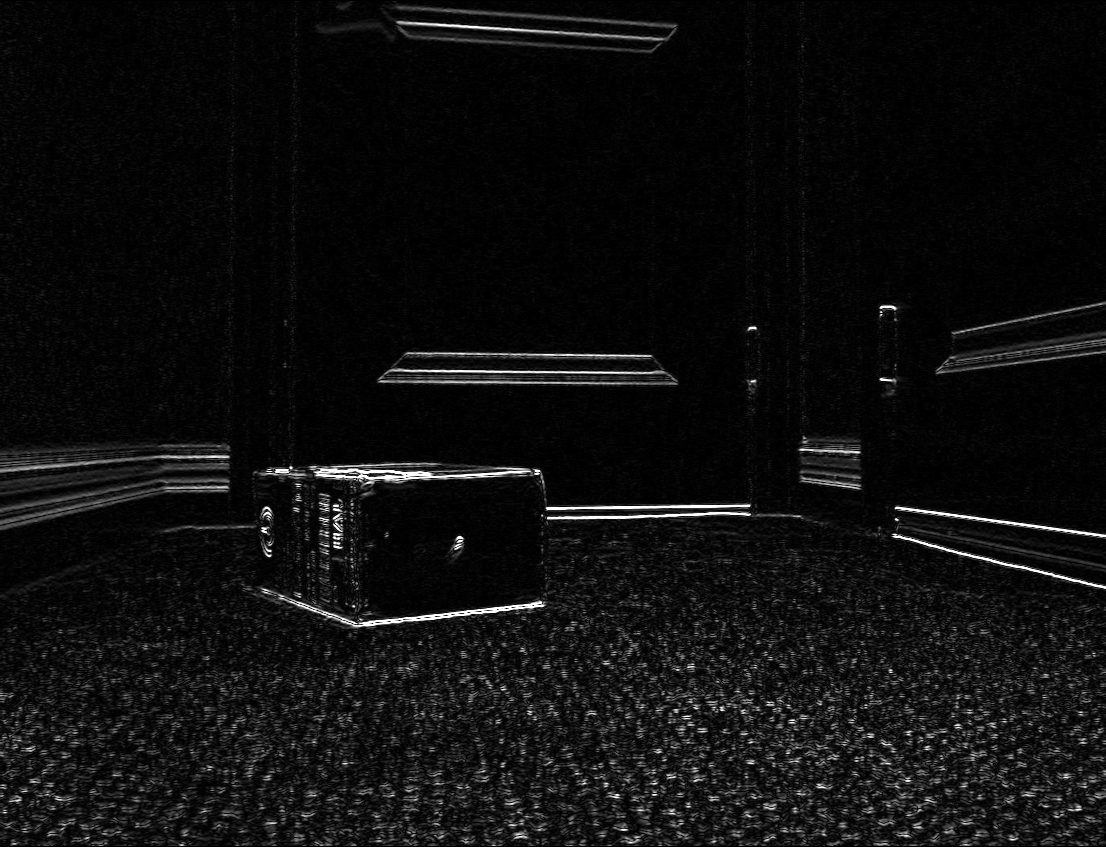}
	  \caption{Sobel Y}
	\end{subfigure}
	\begin{subfigure}[b]{0.45\linewidth}
	  \includegraphics[width=\linewidth]{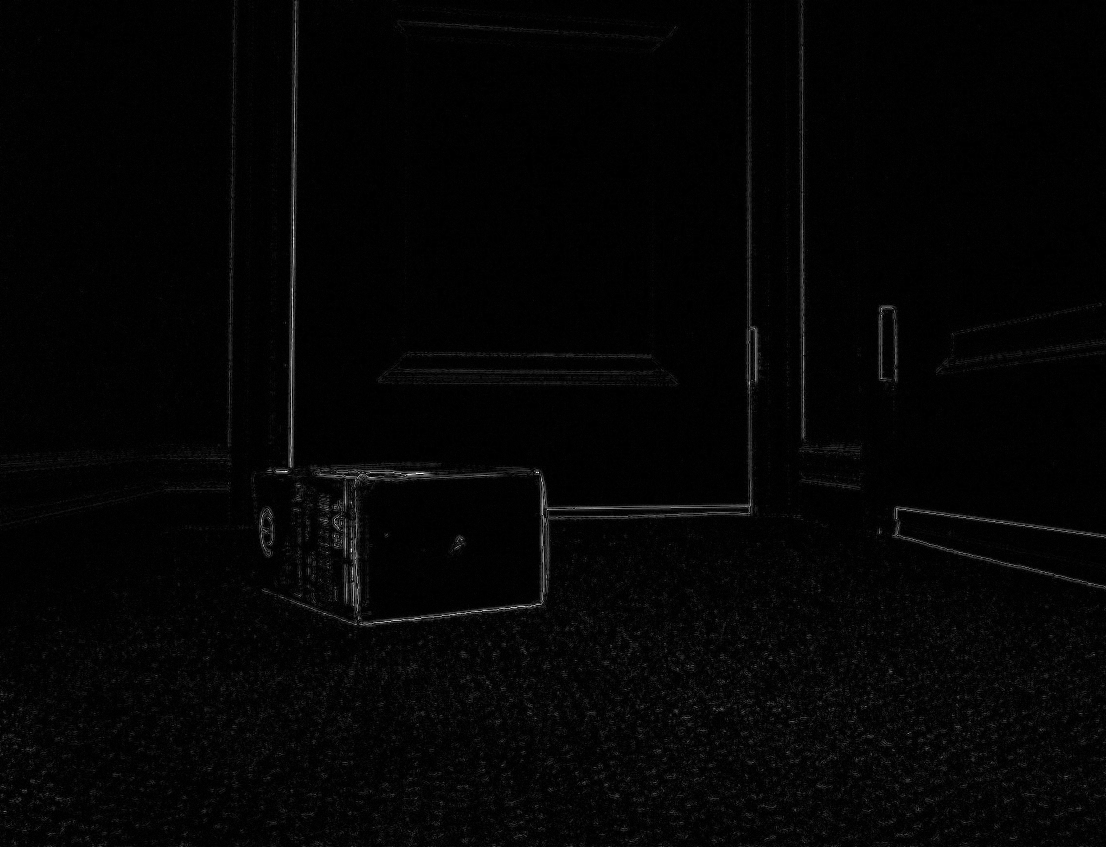}
	  \caption{Laplacian}
	\end{subfigure}
	\caption{The four inputs to the neural network. The main input contains the three colors channels in (a), while the auxiliary input contains (b), (c), and (d)}
	\label{fig:inputs}
\end{figure}

\subsection{Network Architecture}

The goal of semantic segmentation is to classify every pixel in an image into one of a set number of discrete classes \cite{cite:2}. For our use case, there are only two classes: free space and obstacles. Current state-of-the-art models, such as \cite{cite:3} \cite{cite:4}, have millions of parameters that must be trained end-to-end, requiring massive amounts of labeled data and lengthy training time. Another problem with these larger architectures is that they often cannot run fast enough for real-time robot navigation. 

Our specific task of segmenting free space (binary classification) requires much less contextual information than segmenting the multiclass \cite{cite:5} \cite{cite:6} datasets used as benchmarks for development of standard semantic segmentation models. As a result, we can design an architecture much simpler than the state-of-the-art, with the goal of achieving real-time performance on an embedded GPU and efficient training with a small dataset.

We propose an end-to-end semantic segmentation network that can accurately perform pixel-level segmentation of free space. We combine proven techniques from \cite{cite:3} \cite{cite:4} \cite{cite:7} \cite{cite:8} with our own research to create a robust neural network that can run at 55 fps on an embedded NVIDIA Jetson TX2's GPU and can be trained quickly with a only a small labeled dataset. Following is a detailed description of our model architecture. 

\textbf{Note:} All convolutional layers use a 3x3 kernel, stride of 1, dilation rate of 1, and "same" padding (input size divided by stride is same as output size) unless otherwise noted.

\subsubsection{Input}
The main input to the network is a 224x224 RGB image. The frame captured by the camera is undistorted, cropped (only the bottom 2/3 of the image is used because the upper 1/3 will never contain space above the horizon), resized to 224x224, then scaled linearly from values of [0, 255] to [-1.0, 1.0] to prepare it to be used as input to the neural network. \\

\subsubsection{Feature Extractor}
To improve training times, a pretrained classification model of MobileNetV2 \cite{cite:7} (width multiplier = 1.0) is used as the feature extractor for this network. Only the first 13 inverted residual bottleneck blocks from MobileNetV2 are used, with weights initialized by pretraining as on the ImageNet challenge dataset \cite{cite:6}. This is used as a method of transfer learning \cite{cite:27}, which allows the feature extraction capabilities of the model to be learned in a controlled setting, and improves convergence when the full segmentation model is trained. The output of the feature extractor is a 14x14x96 tensor (1/16 of the input size) \\
%TODO inv. res block diagram

\subsubsection{Bottleneck Layers}
Using the remaining bottleneck blocks from MobileNetV2 would reduce the output resolution further, so we replace them with two bottleneck blocks similar to those in \cite{cite:7}, each having an expansion factor of 6 and 160 output filters. The addition of these blocks makes the network deeper and compensates for the layers removed from the end of the feature extractor. Finally, a 1x1 convolution with 256 filters is applied to the output of the second bottleneck layer. \\

\subsubsection{Atrous Convolution Pyramid}
To achieve more precise predictions while reducing the number of operations done by the feature extractor, we use an atrous pyramidal module inspired by \cite{cite:4} \cite{cite:10} to effectively capture information at multiple scales in the feature map. We use four convolutions: one 1x1 (with dilation rate 1) and three 3x3 (with dilation rates {1, 2, 4}), each followed with by a ReLU activation. Batch Normalization \cite{cite:11} was not used here, as the batch size was small, and experiments showed that the network was able to converge successfully without use of Batch Norm. The outputs of these four convolutions are then concatenated along the channel axis, yielding a 14x14x1024 feature map. This is followed by a 1x1 convolution with 256 filters and a ReLU activation, to reduce the depth of the feature map back to 256. \\

\subsubsection{Upsampling}
We use the deconvolution (transpose convolution) layer to upsample the output of the atrous pyramid. The deconvolution has been shown to be extremely successful in upsampling for semantic segmentation in applications such as \cite{cite:8} We use a deconvolution layer with a kernel size of 8x8 and a stride of 4 to upsample the 14x14x256 output of the Atrous Convolution Pyramid to 56x56x256. We find that using deconvolution achieves better accuracy and faster convergence than the nonotrainable bilinear upsampling used in \cite{cite:3}, at the cost of being slightly more computationally intensive. \\
% TODO table of deconvolution

\begin{figure}[!t]
	\centering
	\includegraphics[width=0.9\linewidth]{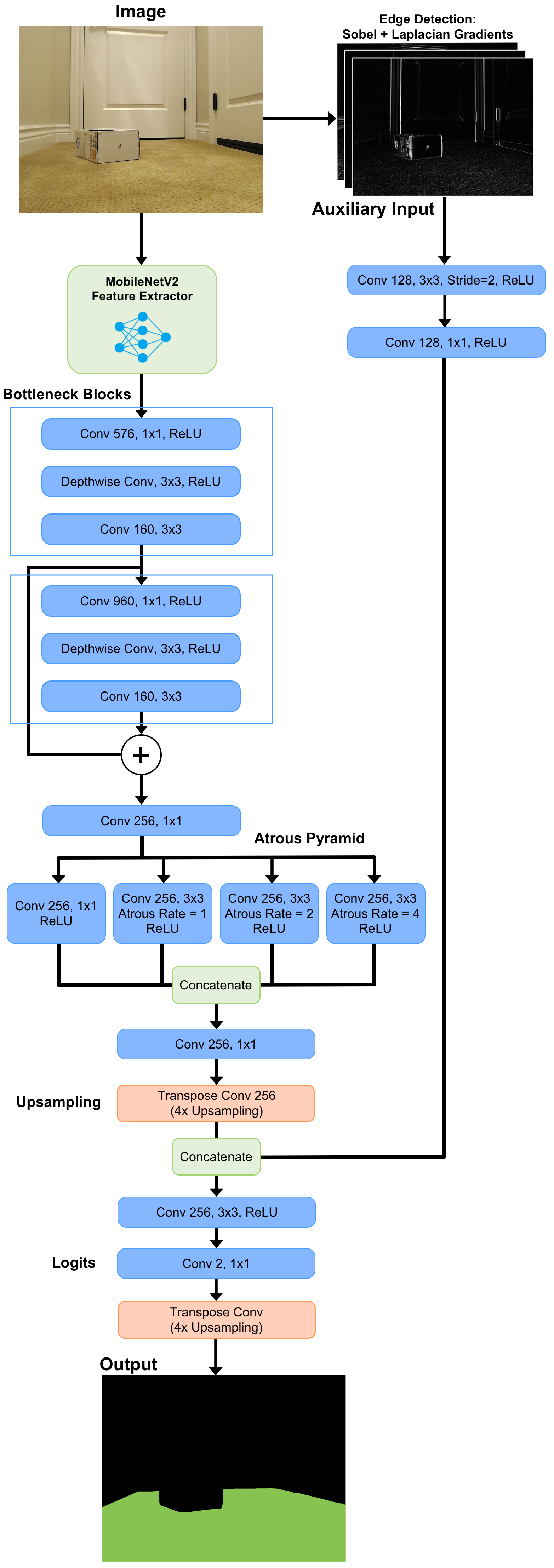}
	\label{fig:arch_diagram}
\end{figure}

\subsubsection{Auxiliary Input}
At this point in the network, the auxiliary input described in the previous section (Input Preprocessing) is added into the network. This input is in the form of a 112x112x3 feature map, scaled to [-1.0, 1.0], and includes the Sobel and Laplacian gradients of the image. The purpose of adding this input is to allow the network to fine-tune its predictions to known edges of objects. The image gradient has been shown \cite{cite:12} \cite{cite:13} to be highly effective in detecting edges of objects in an image. Because the edges of the freespace map (segmentation output) generally correspond with edges of objects, explicitly adding the image gradient at this point helps refine segmentation results at object boundaries. The image gradient is fed through a 3x3 (128 filters) convolution with stride=2, and then a 1x1 (128 filters) convolution with stride=1 to downsample it to 56x56x128. Each convolution includes a ReLU activation. The output of these convolutions is then concatenated to the output of the first part of the network, yielding a 56x56x384 feature map. \\
  
\subsubsection{Logits and Output}
Finally, to create the output logits, this feature map is fed through a 3x3 convolution with 256 filters and ReLU activation, followed by a 1x1 convolution with 2 filters, which forms the output logits layer for the network. Finally, similarly to \cite{cite:3}, the logits are upsampled, again using a deconvolution layer. The output of this upsampling has a shape of 224x224x2, and is fed through a softmax activation to obtain a probability vector at each point on the image. In this vector, channel 0 represents obstacles and background while channel 1 represents free space. To create a single freespace map, the channel with the higher probability at each point is used as the segmentation output, yielding a 224x224 image of binary values (0 = background, 1 = free space). \\

\begin{figure}[h!]
	\centering
	\includegraphics[width=\linewidth]{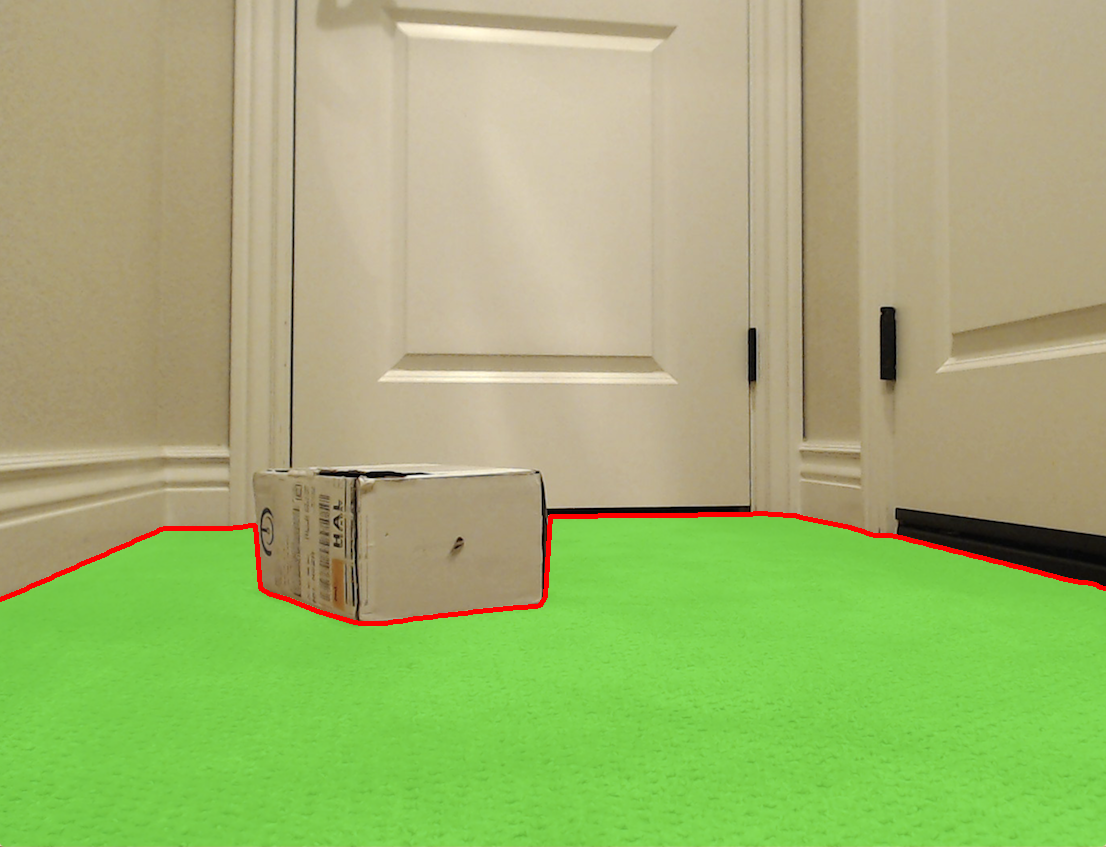}
	\caption{The freespace map produced by the neural network, superimposed on the input image.}
	\label{fig:output}
\end{figure}

\subsection{Inference Techniques}
Freespace segmentation must be run in real-time on a low-cost embedded GPU to be useful for indoor robot pathfinding.

Real-time Inference: We use an NVIDIA Jetson TX2 SoC as our reference platform. It is low-cost, has a low power draw (15W), and has a small size and rugged carrier board, making it an ideal processing platform for the types of indoor and outdoor robots that our research targets. The goal of this neural network is to be able to run in real-time, which we define as at least 15fps, on a 200x200 or greater resolution input image.

TensorRT Optimization: The Jetson TX2, optimized for deep learning, includes hardware-level support for TensorRT, which can massively accelerate neural network performance. We are also able to use FP16 (16-bit floating point) compute, which halves the memory bandwidth required for inference, improving the runtime significantly. We take advantage of TensorRT to accelerate our network to maximum possible performance. In order to support TensorRT, all Relu6 operations (ReLU layers with a maximum value of 6.0) are replaced with the following expression: $$\text{ReLU}(x) - \text{ReLU}(x - 6)$$ which is equivalent but supported natively by TensorRT. The use of TensorRT nearly triples our inference performance, which is a critical improvement for real-time use on a robot. \\

\subsection{Navigation Stack Integration}

To effectively use the freespace map outputted by the semantic segmentation network, it must be converted to a format usable by pathfinding and navigation algorithms to control a robot. We use ROS (Robot Operating System) \cite{cite:14} because it has SLAM mapping and localization capabilities built-in, as well as the ability to use arbitrary 3D pointclouds as input to the navigation algorithms \cite{cite:15}. We set up ROS and use its navigation stack to enable basic LIDAR-driven navigation capabilities on the robot. We then integrate our segmentation output into the costmap, which tracks obstacles and free space. In order to convert the 2D camera-perspective freespace map into a 3D pointcloud, we must use the following process:

\begin{figure}[h!]
	\centering
	\includegraphics[width=\linewidth]{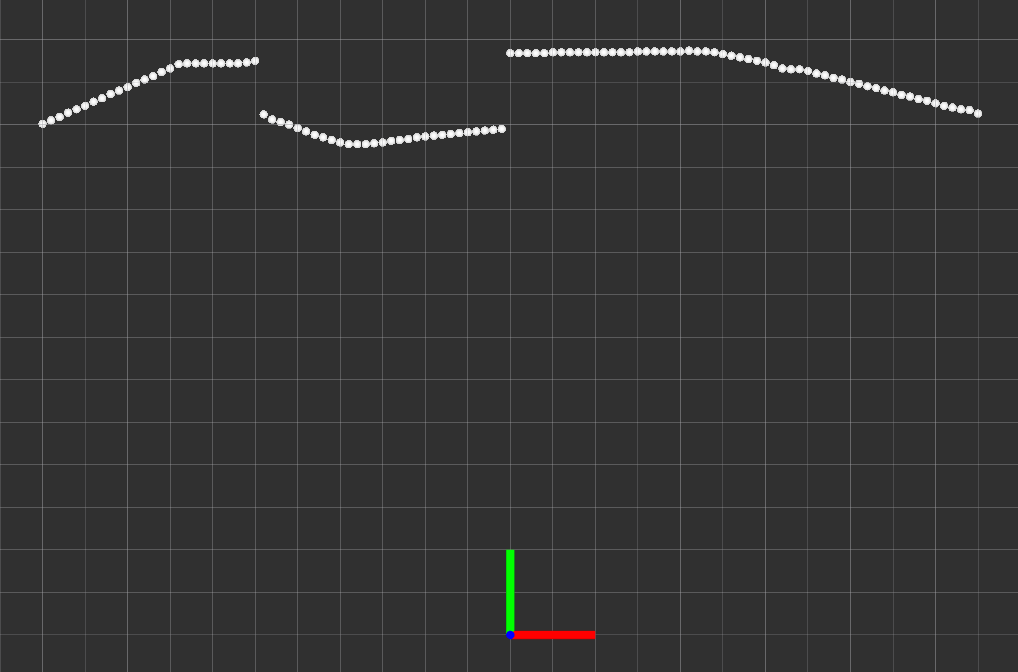} % TODO get better pcl if possible
	\caption{The 3D pointcloud produced from the freespace map and used for real-time navigation.}
	\label{fig:output}
\end{figure}

\subsubsection{Freespace Map Border Extraction}
The output freespace map is a full-resolution image with many sharp edges and a large number of border points (border points are defined as points which lie along the edge between free space and obstacles). We propose three methods of extracting the most important border points from the image, and demonstrate optimal and sub-optimal cases for each:

\setlength{\fboxsep}{0pt} % Border around image touching image
\setlength{\fboxrule}{1pt} % Border around image = 1pt

\begin{enumerate}[label=(\roman*)]
	\item Vertical Line Projection: Vertical lines are drawn uniformly across the image. The lowest (closest to the camera) border point which intersects with each line is added to the pointcloud. The pointcloud density can be varied by the number of values of $n$ to change the number of lines. This projection works well in many cases, but produces less precise results when an obstacle border is between two vertical lines, which causes the line to miss most of the points on the border. \\
	For integer values $n$ distributed uniformly in the interval $[0, w)$, we define the projected line as:
	\[a_n = \begin{cases} x=n \\ y=t \end{cases} t\in [0, h) \cap \mathbb{Z}\]
	\begin{figure}[h!]
		\centering
		\begin{subfigure}[b]{0.475\linewidth}
			\fbox{\includegraphics[width=\linewidth]{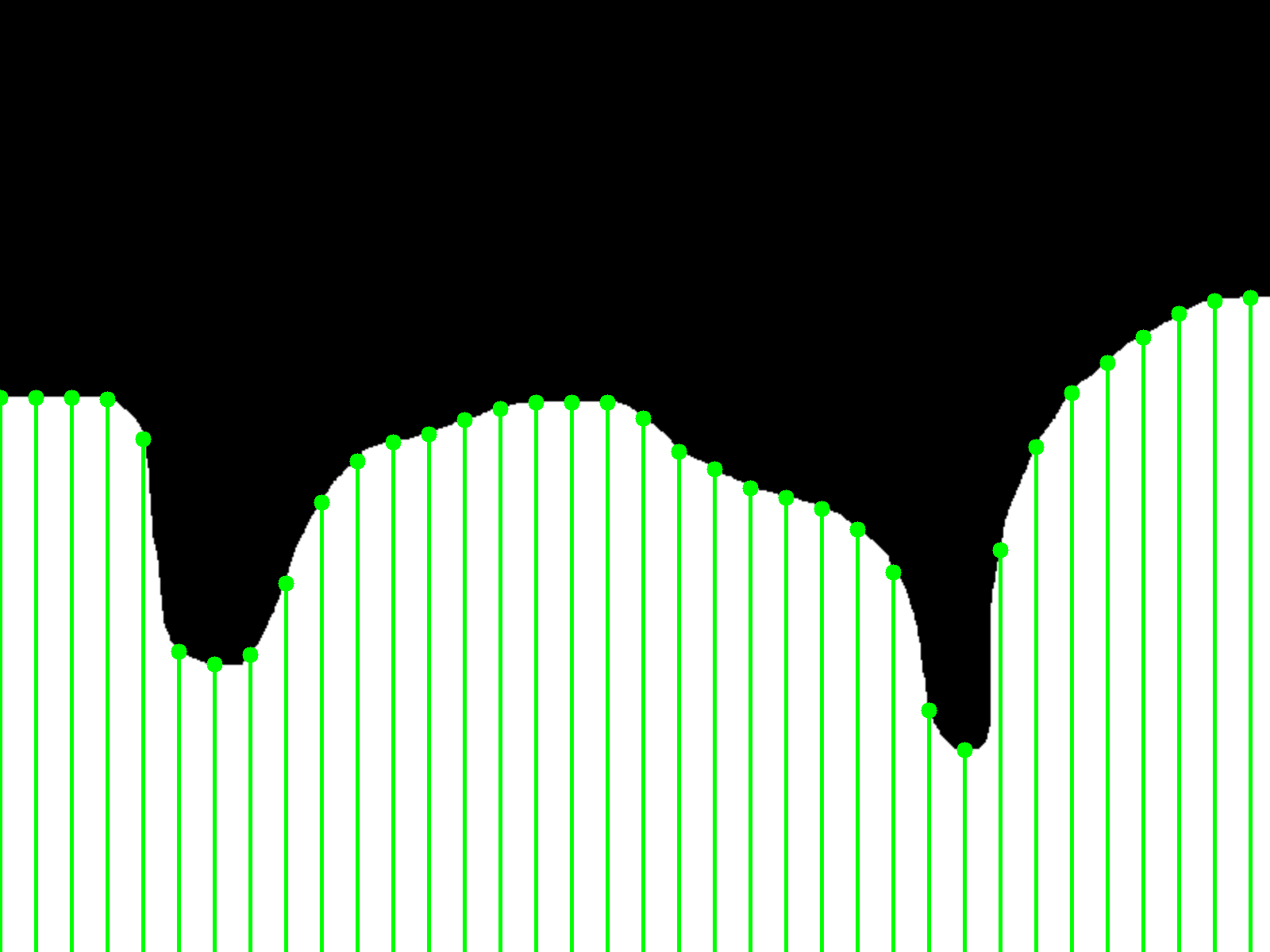}}
			\caption{Optimal Case}
		\end{subfigure}
		\begin{subfigure}[b]{0.05\linewidth} \end{subfigure}
		\begin{subfigure}[b]{0.475\linewidth}
			\fbox{\includegraphics[width=\linewidth]{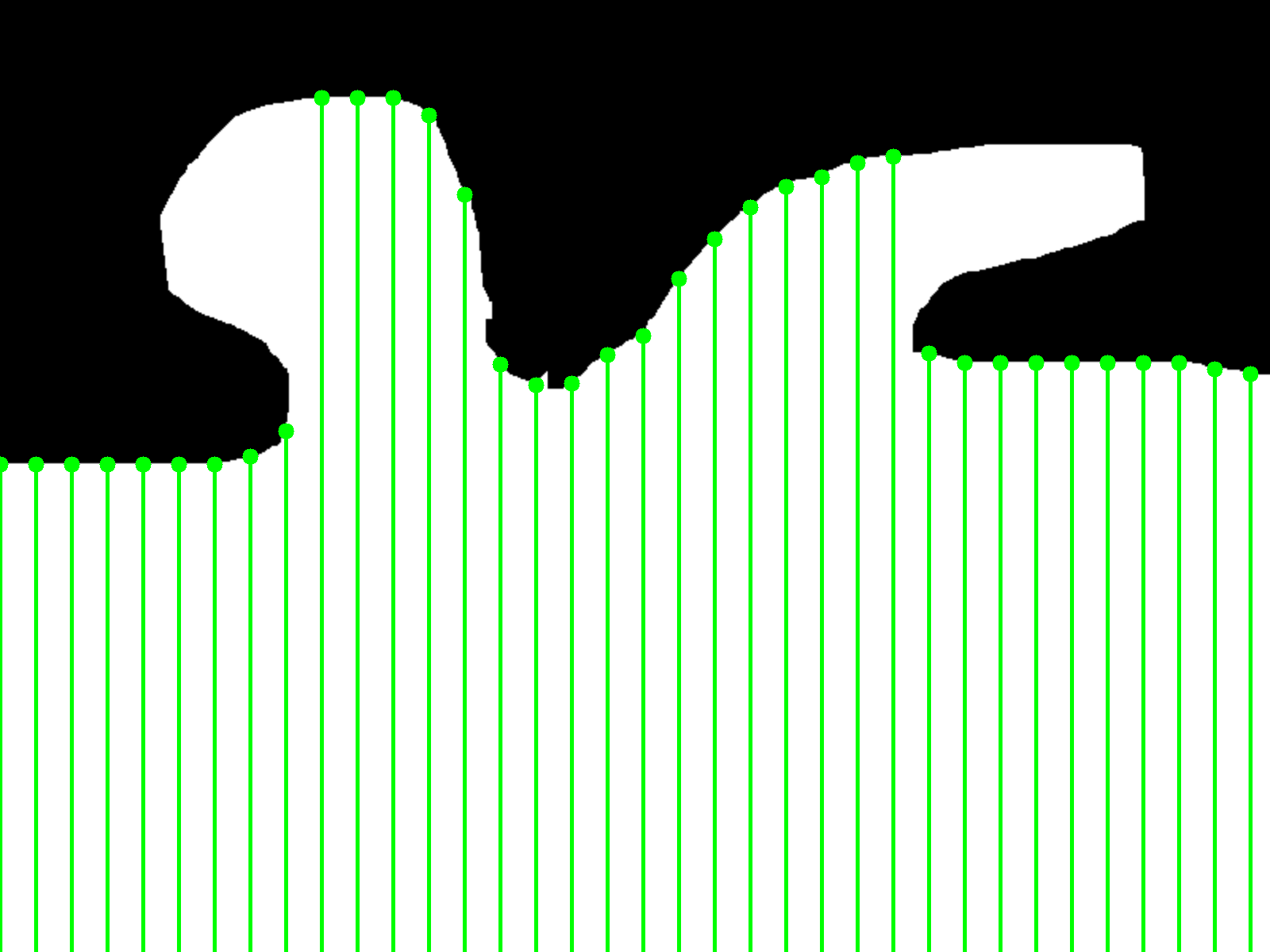}}
			\caption{Sub-Optimal Case}
		\end{subfigure}
		\caption{Examples of vertical line projection. Note how the selected points accurately represent the border in figure (a), while they do not capture two regions in figure (b).}
		\label{fig:vertical_proj}
	\end{figure}
	\item Polar Projection: Lines are drawn from the point at the bottom-center of the image ($\frac{w}{2}$) extending outward. These lines are drawn at uniform angle increments, based on the equation below. The lowest border point that intersects each line is added to the pointcloud. The pointcloud density can be varied by changing the number of values $n$. This projection works well in many cases, including some of the vertical line projection's failure cases, but fails when an obstacle (not close to the center or edge of the image) is extending towards the robot. \\
	For integer values $n$ distributed uniformly in the interval $[0, \pi])$, we define the projected line as:
	\[a_n = \begin{cases} x = \floor{\frac{w}{2} - t \cos{n}} \\ y = \floor{t \sin{n}} \end{cases}t\in [0, \text{min}(\frac{w}{2 \cos{n}}, \frac{h}{\sin{n}}))\]
	\begin{figure}[h!]
		\centering
		\begin{subfigure}[t]{0.475\linewidth}
			\fbox{\includegraphics[width=\linewidth]{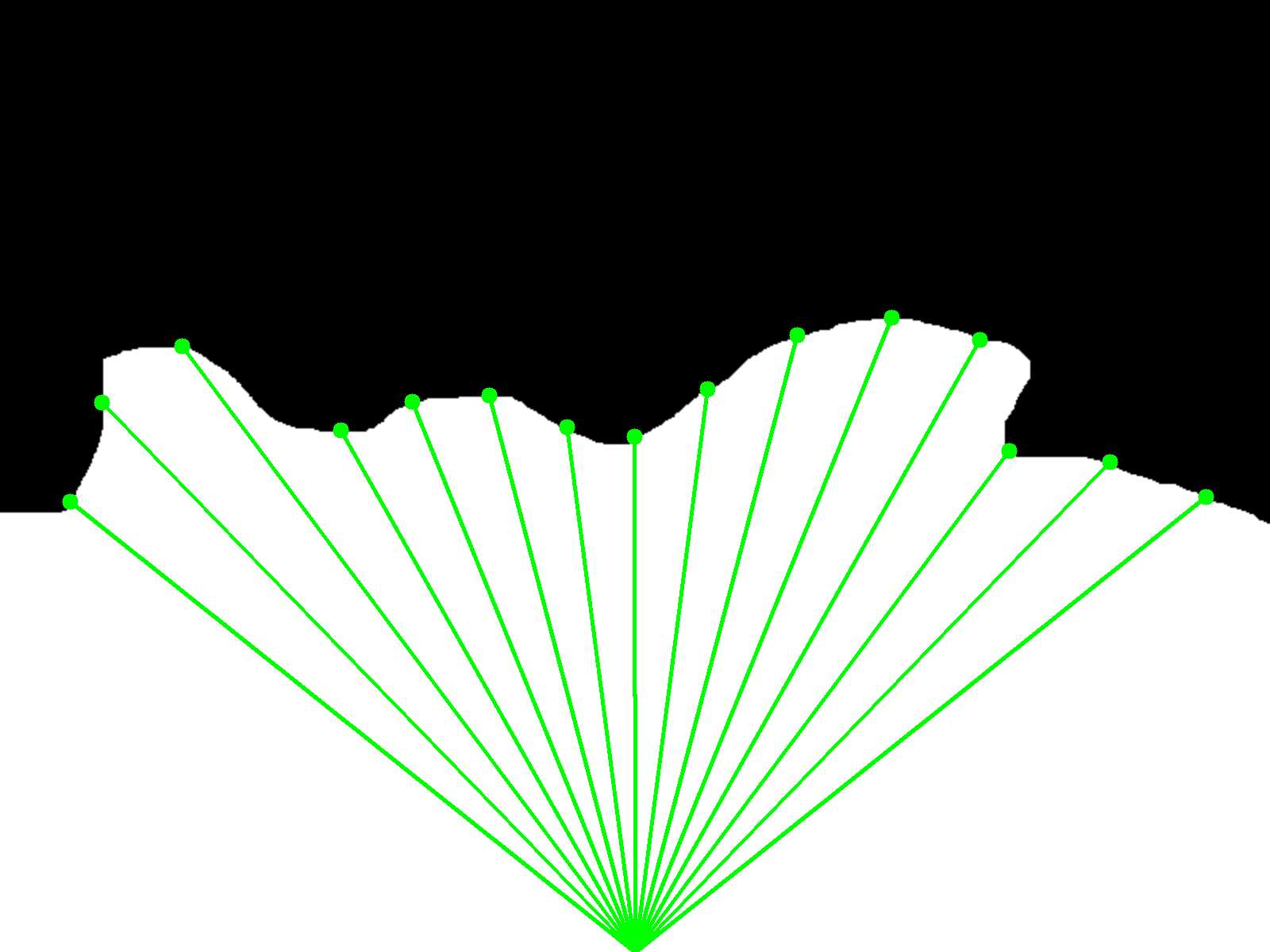}}
			\caption{Optimal Case}
		\end{subfigure}
		\begin{subfigure}[t]{0.05\linewidth} \end{subfigure}
		\begin{subfigure}[t]{0.475\linewidth}
			\fbox{\includegraphics[width=\linewidth]{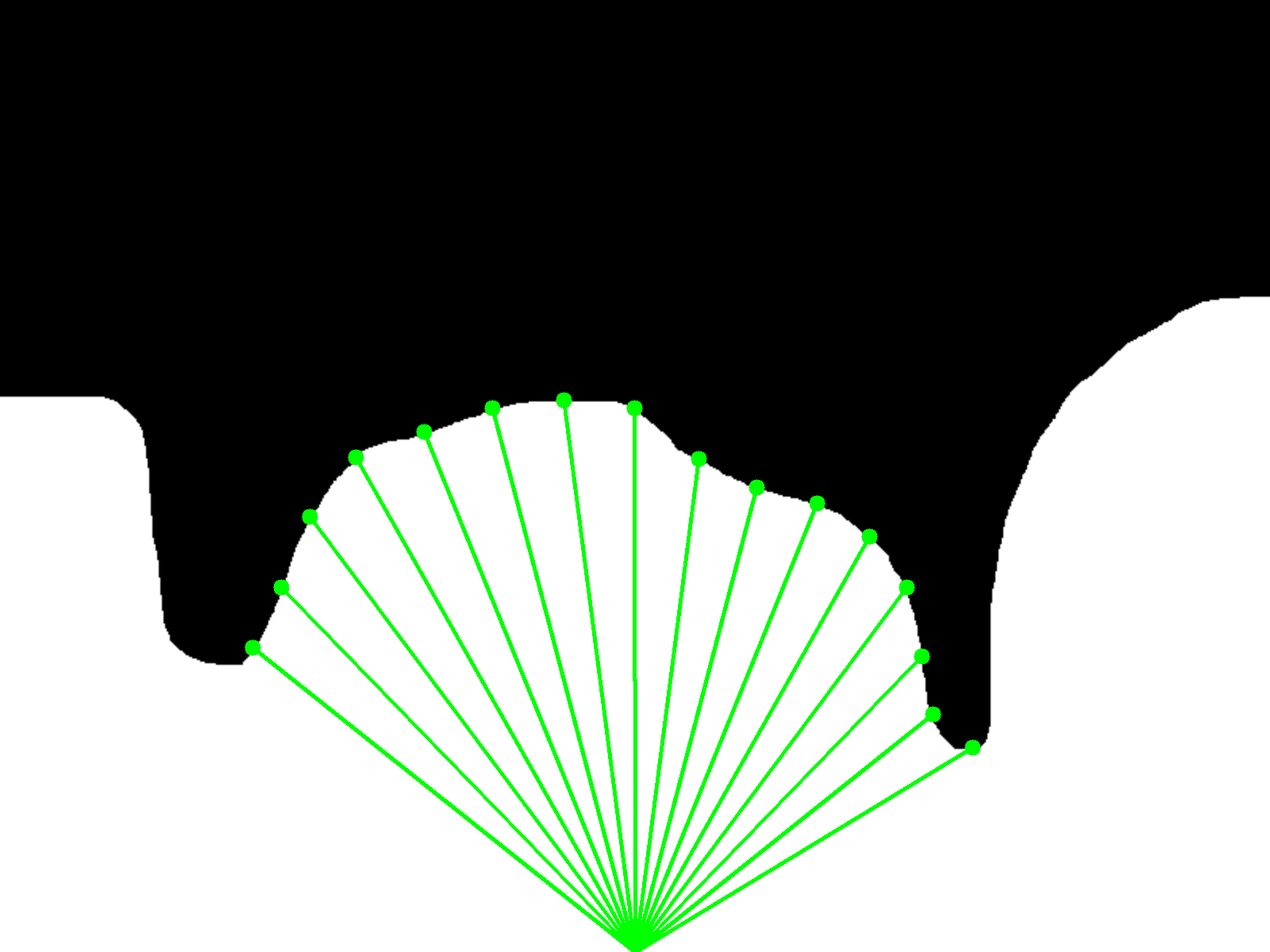}}
			\caption{Sub-Optimal Case}
		\end{subfigure}
		\caption{Examples of polar projection. Note how the selected points cover the entire border in figure (a), but are cut off by the lowest obstacle points in figure (b).}
		\label{fig:polar_proj}
	\end{figure}
	\item Contour Extraction: Instead of using lines drawn from the bottom of the image, contours are extracted in the freespace map using methods from 17. Points along these contours are then used for the pointcloud. The density of the pointcloud can be controlled by selecting less points from the contours. This projection works well in most cases, but may produce improper pointclouds in cases where free space is visible both in front of and behind an obstacle.
	\\
	\begin{figure}[h!]
		\centering
		\begin{subfigure}[t]{0.475\linewidth}
			\fbox{\includegraphics[width=\linewidth]{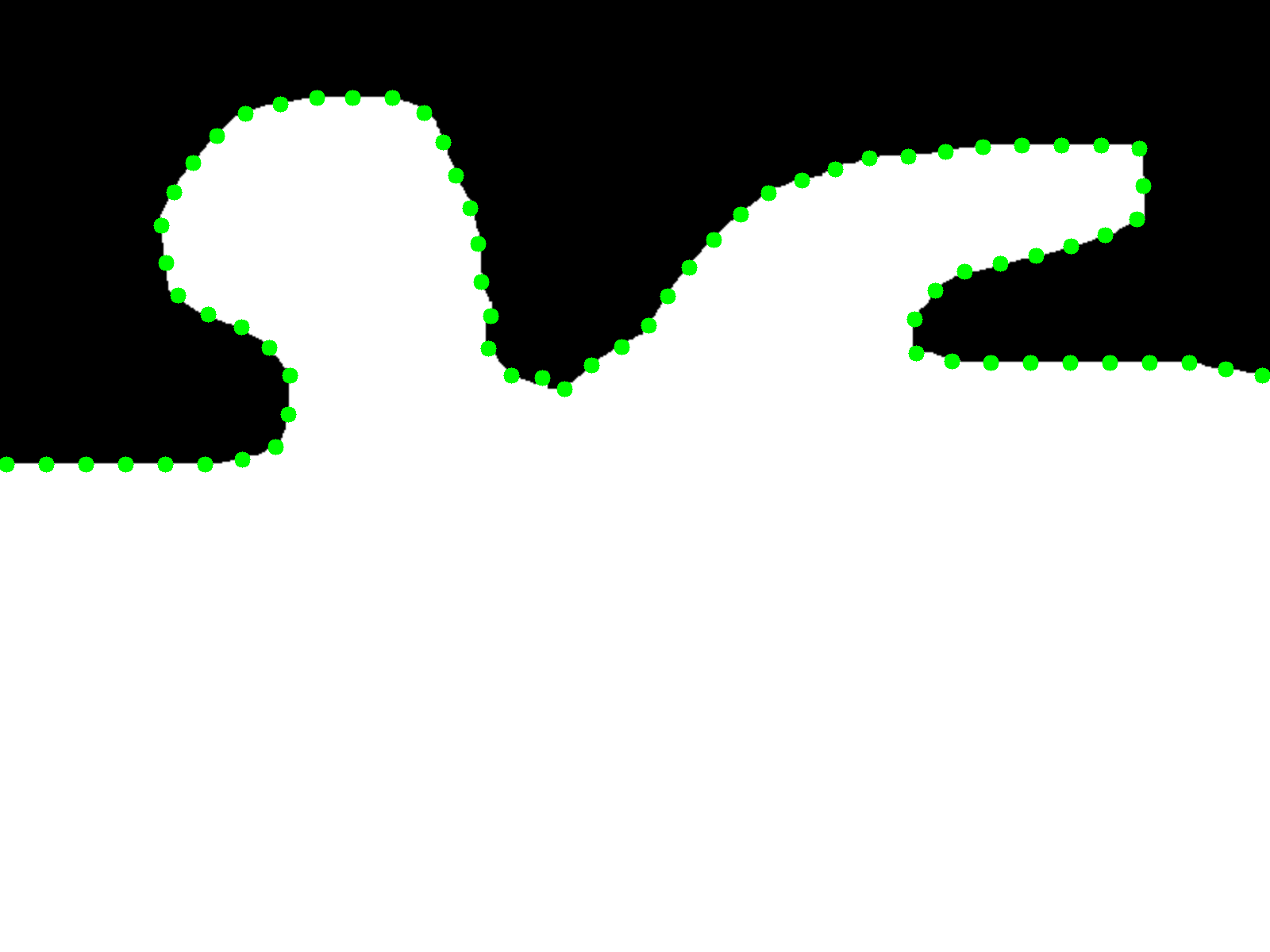}}
			\caption{Optimal Case}
		\end{subfigure}
		\begin{subfigure}[t]{0.05\linewidth} \end{subfigure}
		\begin{subfigure}[t]{0.475\linewidth}
			\fbox{\includegraphics[width=\linewidth]{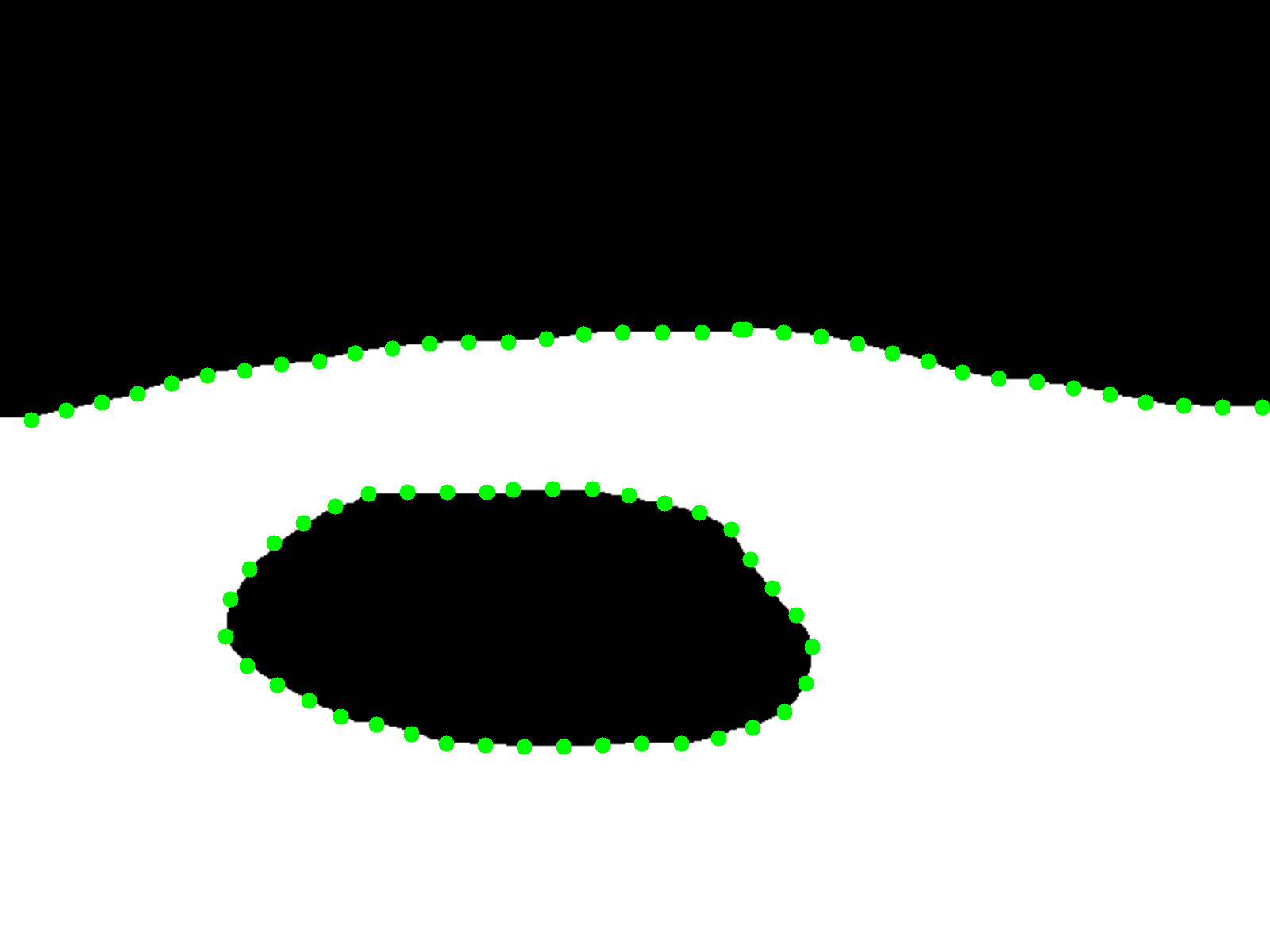}}
			\caption{Sub-Optimal Case}
		\end{subfigure}
		\caption{Examples of contour extraction. Note how the contour points accurately match the borders in both images, but they may create a self-overlapping pointcloud in situations such as (b).}
		\label{fig:polar_proj}
	\end{figure}
\end{enumerate}

We find that each of these methods is useful in different scenarios, with the polar projection being the most successful in our trials with a navigation stack, most likely as a result of how it replicates the structure of commonly-used LIDAR data. \\

\subsubsection{3D Pointcloud Creation}
After extracting the border points (where free space ends) on the 2D freespace map, these points must be projected to a 3D pointcloud in order to use them for robot navigation. We use the pinhole camera model to transform all 2D image-space points into 3D world-space points. By assuming that all border points are at ground level, we are able to determine the positions of all points in 3D and create a 3D pointcloud. These 3D points are then added to a ROS pointcloud data packet and added to the navigation costmap. This costmap, created by combining pointclouds and robot odometry, can be used for safe robot navigation and obstacle avoidance. 

$$
s \begin{bmatrix} u \\ v \\ 1 \end{bmatrix} = \begin{bmatrix} f_x & 0 & c_x \\ 0 & f_y & c_y \\ 0 & 0 & 1 \end{bmatrix} \begin{bmatrix} r_{11} & r_{12} & r_{13} \\ r_{21} & r_{22} & r_{23} \\ r_{31} & r_{32} & r_{33} \end{bmatrix} \begin{bmatrix} X \\ Y \\ Z \end{bmatrix} 
$$

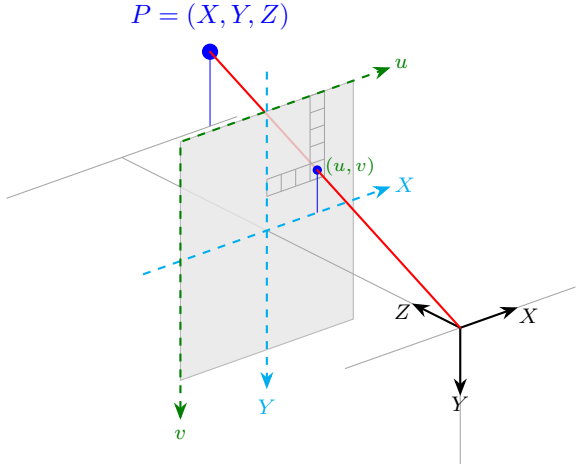
\begin{figure}[h!]
	\centering
	\tdplotsetmaincoords{-65}{-40}
	\begin{tikzpicture}
	[
		tdplot_main_coords,
		>=Stealth,
		my dashed/.style={dashed, thick, ->, shorten >=-15pt, shorten <=-15pt, every node/.append style={font=\footnotesize}},
		my box/.style={thin, gray!70},
		my blue/.style={line width=0, blue, every node/.append style={fill, blue, circle, inner sep=0pt, minimum size=#1*3.5pt, anchor=center, outer sep=0pt}},
		my label/.append style={midway, font=\scriptsize},
		my vectors/.style={green!50!black, {Stealth[scale=.75]}-{Stealth[scale=.75]}},
		my red/.style={thick, red, line cap=round},
		my grey/.style={gray!70},
		description/.style={draw=gray!70, thick, line cap=round, every node/.style={align=center, font=\scriptsize\sffamily, anchor=north}},
	]
	\draw [my grey] (0,4,0) -- (0,7,0) (-2,7,0) -- (2,7,0);
	\coordinate (o) at (0,0,0);

	\path [draw=gray!70, text=gray, fill=gray!20, opacity=0.8, text opacity=1] (-1.5,4,1.75) coordinate (a) -- ++(0,0,-3.5) coordinate (b) -- ++(3,0,0) coordinate (c) -- ++(0,0,3.5) coordinate (d) -- cycle node [pos=.95, above, sloped, anchor=south west] {};

	\draw [my grey] (-2,0,0) -- (2,0,0) (0,0,0) -- (0,4,0) (0,0,0) -- (0,0,2);
	\draw [thick, ->, every node/.style={font=\footnotesize, inner sep=0pt}] (o) node [anchor=north west] {} (o) edge node [pos=1, anchor=north east] {$Z$} ++(0,1,0) edge node [pos=1, anchor=north] {$Y$} ++(0,0,1) -- ++(1,0,0) node [anchor=north west] {$X$};
	\draw [my box] (o) ++(0,4,-.5) coordinate (p1) -- ++(1,0,0) coordinate (p2) -- ++(0,0,-1.25) coordinate (p3);
	\foreach \i in {0,1,...,4} \draw [my box] (p1) ++(\i*.25,0,0) -- ++(0,0,-.25);
	\foreach \i in {0,1,...,5} \draw [my box] (p2) ++(0,0,-\i*.25) -- ++(-.25,0,0);
	\draw [my box] (p1) ++(0,0,-.25) -- ++(.75,0,0) -- ++(0,0,-1);
	\draw [my dashed, cyan] ($(b)!1/2!(c)$) -- ($(d)!1/2!(a)$) node [below=15pt, anchor=north] {$Y$};
	\draw [my dashed, cyan] ($(b)!1/2!(a)$) -- ($(d)!1/2!(c)$) node [above right=17pt, anchor=north west] {$X$};
	\draw [my dashed, green!50!black, <->] (a) node [below=15pt, anchor=north] {$v$} -- (b) -- (c) node [above right=17pt, anchor=north west] {$u$};
	\path [green!50!black, every node/.style={font=\scriptsize, inner sep=0pt}] (p2) node [above right, anchor=south west] {$(u,v)$};
	\path (p2) ++(-.125,0,0) coordinate (q2) ++(0,0,-.125) coordinate (r2);
	\draw [my blue=1] ($(0,4,0)+($(q2)-(p1)$)$) coordinate (s2) -- (r2) node (d1) {};
	\scoped[on background layer]{\draw [my blue=1.75] ($($1.75*($(s2)-(0,4,0)$)$)+(0,7,0)$) -- ++($1.75*($(r2)-(s2)$)$) node (d2) [label={[label distance=-20pt]above:{$P=(X,Y,Z)$}}] {};}
	\draw [my red] (o) -- (d1.center);
	\scoped[on background layer]{\draw [my red] (d1.center) -- (d2.center);}
	\end{tikzpicture}
	\caption{A visualization of the pinhole camera model used to convert 2D image points to 3D pointcloud points.}
	\label{fig:3dpclmodel}
\end{figure}

\subsubsection{Blur Detection}
When a robot is moving at a higher velocity, the image produced by the camera is frequently blurred. We find that the output of the semantic segmentation is less accurate when using a blurred image. In order to prioritize less blurred (more accurate segmentation) images, we calculate the blur factor of the image and add this value to the pointcloud, such that the less blurred images have more effect on the final navigation costmap. To calculate the blur of an image we use the method proposed in \cite{cite:18}: the variance of the discrete Laplacian (second derivative gradient) of the image. The following equation shows how the blur factor $\beta$ is calculated, given that the function $L(x, y)$ is a discrete convolution of the input image $I(x, y)$ where $x$ and $y$ are image coordinates:
\begin{align*} 
	L(x, y) &= \begin{pmatrix} \phantom{-}0 & -1 & \phantom{-}0 \\ -1 & \phantom{-}4 & -1 \\ \phantom{-}0 & -1 & \phantom{-}0 \end{pmatrix} \ast I \\
	\overline{L} &= \frac{1}{XY}\sum\limits_{x}^{X} \sum\limits_{y}^{Y} | L(x, y) | \\
	\beta = \sigma^2 &= \sum\limits_{x}^{X} \sum\limits_{y}^{Y} (|L(x, y)| - \overline{L})^2 \\
\end{align*}

The scaling of $\beta$ can be determined experimentally based on the camera intrinsics and image resolution. This $\beta$ value can then be added as an intensity parameter in the pointcloud data packet.

\section{Experimental Evaluation}
We implemented the entire approach described in this paper using the Google TensorFlow library, and OpenCV for preprocessing input data. Additionally, we use NVIDIA TensorRT to accelerate neural network inference on embedded platforms. We then test our approach using an iRobot Create 2 indoor robot with an RGB camera and NVIDIA Jetson TX2.

\subsection{Training}

\subsubsection{Data Augmentation}
In order to train the neural network to generalize from only a small training dataset, we apply data augmentation on the training data. We apply two types of augmentations to improve the generalization of the network. First, we randomly apply a small blur and/or brightness change to some of the images. This improves the network's performance on blurry images (which are often captured if the robot is moving at a high velocity), as well as making it more robust to different lighting conditions. Second, we apply affine transforms to the images. We apply random scaling, horizontal flip, translation, rotation, and shear transforms to the image. We fill edges of the transformed images by reflecting the original image, such that the image appears continuous instead of having sharp edges.
When training, we use a generator-based approach. For each batch, images are randomly selected from the training dataset. These images (and their corresponding labels) are augmented randomly using our augmentation pipeline. This allows the generation of an infinite amount of randomly-augmented data, preventing the network from overfitting to the small dataset. \\

\subsubsection{Training Setup}
We train our neural network using Stochastic Gradient Descent using the Adam optimizer with binary crossentropy as a loss function. We use a variation of the "poly" learning rate decay policy proposed in \cite{cite:19}: $$LR(\text{epoch}) = 0.0006 \cdot \left(\frac{1000 - \text{epoch}}{1000}\right)^{0.9}$$
The network is trained for 1000 epochs, after which we find no significant improvement in accuracy. Training takes about 15 hours on an NVIDIA GTX 1080Ti. To evaluate accuracy of the network output, we use the mIoU metric (mean intersection-over-union) averaged across our validation dataset.

\subsection{Performance Evaluation}
We evaluate performance using the mIOU metric with various configurations of our network and compare. We find that, although removing the atrous pyramid and/or auxiliary input reduces the number of FLOPS, it also lowers the accuracy. Our optimal performance is about 94.9\% mIOU with all layers of the neural network enabled. We find that this is enough accuracy for real-time navigation. Examples of the network output are in Appendix A.

\subsection{Runtime Evaluation}
We evaluate our neural network performance on a variety of computing platforms that can be used on embedded robots. We use the NVIDIA Jetson TX2 as our reference platform, and an NVIDIA GTX 1080Ti as a second testing platform (which can be used on larger robots such as those used outdoors). Note: all tests on the TX2 are done after setting it to MAX-N mode, which increases its max TDP to 15W and increases core clock speeds.

Our model runs at 18.9 fps on the Jetson TX2's GPU (standard TensorFlow). Optimizing with TensorRT nearly triples the network's performance to 55.6 fps. If the GPU is being used for other purposes (stereo matching, path generation, other neural networks, etc.), our network can run on the TX2's CPU at 3.6 fps, which is still viable for real-time navigation on slower-moving robots. If more precise segmentation is desired, our network can run with a 512x512 input size at 13.0 fps on the TX2's GPU. We also tested at higher resolutions using a GTX 1080Ti GPU, which can be used on larger robots.

\begin{table}[h!]
	\centering
	\renewcommand*{\arraystretch}{1.4}
	\begin{tabular}{l||rr||rr}
	\hline
	\multirow{2}{*}{Inference Platform}    & \multicolumn{2}{c||}{FP32} & \multicolumn{2}{c}{FP16} \\
						  & ms          & fps         & ms          & fps          \\
	\hline
	Jetson TX2 (TensorRT) & 24          & 41.7        & 18          & 55.6         \\
	Jetson TX2 (GPU)      & 63          & 15.9        & 57          & 17.5         \\
	Jetson TX2 (CPU)      & 277         & 3.6         & ---         & ---          \\
	GTX 1080Ti            & 8           & 125.0       & 12          & 83.3         \\
	\hline 
	\end{tabular}
	\caption{Runtime speed of the neural network at 224x224 resolution on different platforms}
	\label{table:speed}
\end{table}

\begin{table}[h!]
	\centering
	\renewcommand*{\arraystretch}{1.4}
	\begin{tabular}{l||rr||rr||rr}
	\hline
	\multirow{2}{*}{Inference Platform}    & \multicolumn{2}{c||}{224x224} & \multicolumn{2}{c||}{512x512} & \multicolumn{2}{c}{768x768} \\
						  & ms          & fps         & ms          & fps          & ms          & fps          \\
	\hline
	TX2 (TensorRT) FP16 & 18          & 55.6        & 77          & 13.0         & 170          & 5.9         \\
	GTX 1080Ti FP32           & 8           & 125.0       & 21          & 47.6         & 38          & 26.3         \\
	\hline 
	\end{tabular}
	\caption{Runtime speed of the neural network at different resolutions}
	\label{table:speed2}
\end{table}

\section{Conclusion}
In this paper, we presented an approach for detection of safe terrain on a robot solely from RGB camera data. We build and train a deep CNN that takes advantage of residual blocks, atrous convolutions, and specialized preprocessing to perform pixel-wise semantic segmentation of freespace in an image. The addition of auxiliary edge-detection inputs allows the network to perform better and makes it easier to train. We also successfully integrate the output of the neural network into a real-time navigation stack, allowing it to be used for robot pathfinding and obstacle avoidance. Our results show that the system generalizes well, is suitable for real-time operation, and runs at around 55fps on a low-power embedded GPU.

\bibliographystyle{IEEEtran}
\bibliography{IEEEfull,bibliography}

\newpage

\section*{Appendix A: Qualitative Evaluation}

\doubleimage{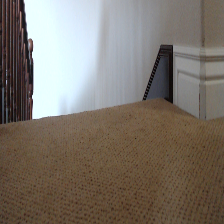}{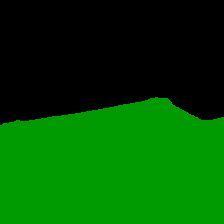} 
\doubleimage{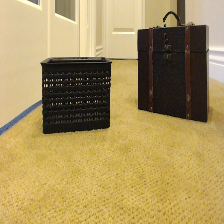}{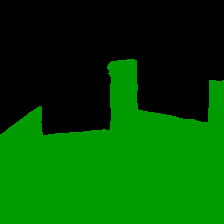}
\doubleimage{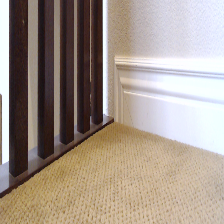}{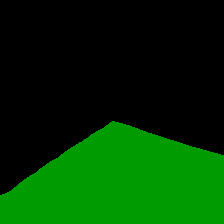}
\doubleimage{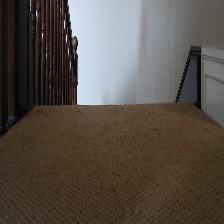}{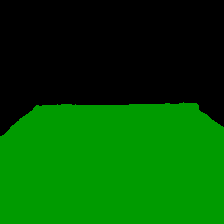}
\doubleimage{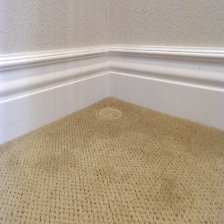}{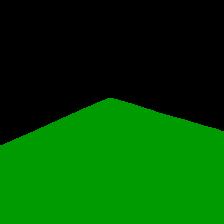}
\doubleimage{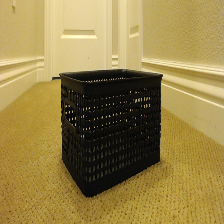}{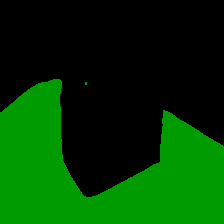}
\doubleimage{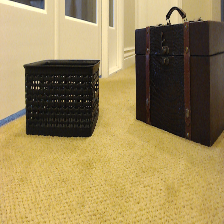}{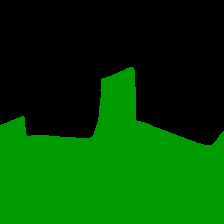}
\doubleimage{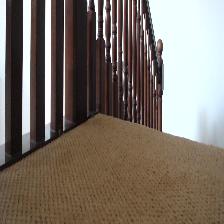}{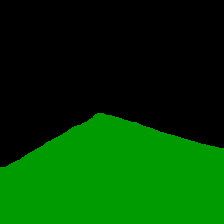}
\doubleimage{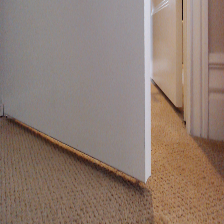}{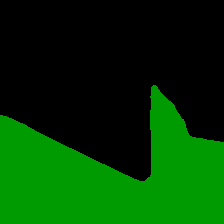}
\doubleimage{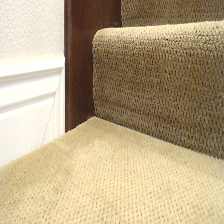}{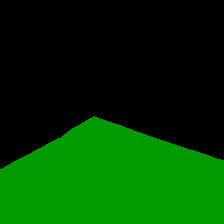}
\doubleimage{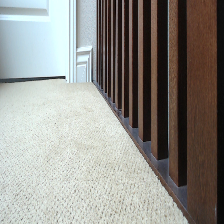}{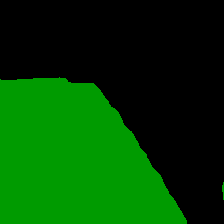}
\doubleimage{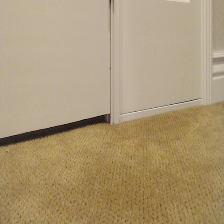}{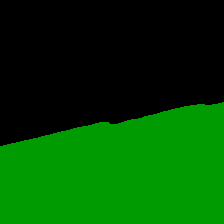}
\doubleimage{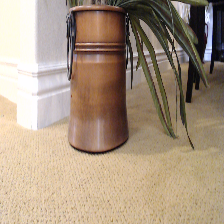}{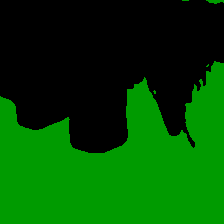}
\doubleimage{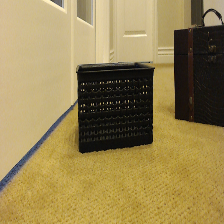}{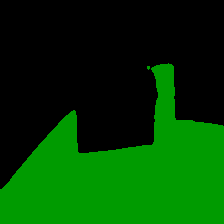}

\end{document}